# A Method for Integrating Utility Analysis into an Expert System for Design Evaluation under Uncertainty

Deborah L. Thurston*
Yun Qi Tian
Decision Systems Laboratory
Department of General Engineering
University of Illinois at Urbana-Champaign

## Abstract

In mechanical design, there is often unavoidable uncertainty in estimates of design performance. Evaluation of design alternatives requires consideration of the impact of this uncertainty. Expert heuristics embody assumptions regarding the designer's attitude towards risk and uncertainty that might be reasonable in most cases but inaccurate in others. We present a technique to allow designers to incorporate their own unique attitude towards uncertainty as opposed to those assumed by the domain expert's rules. The general approach is to eliminate aspects of heuristic rules which directly or indirectly include assumptions regarding the user's attitude towards risk, and replace them with explicit, user-specified probabilistic multiattribute utility and probability distribution functions. We illustrate the method in a system for material selection for automobile bumpers.

## 1. INTRODUCTION

Design evaluation requires simultaneous consideration of several attributes under uncertainty, such as manufacturing cost. At the preliminary design stage, there is frequently a great deal of uncertainty in the estimates of performance levels. In later stages of design, the degree of uncertainty often decreases but might not diminish entirely. This uncertainty has a detrimental impact on the desirability of design alternatives.

* Send all correspondence to: Deborah Thurston
Department of General Engineering, 104 S. Mathews Ave., University of Illinois at Urbana-Champaign, Urbana, IL, 61801.
telephone: 216-333-6456,
e-mail: thurston@uxh.cso.uiuc.edu

### 1.1. EXPERT SYSTEMS FOR MECHANICAL ENGINEERING DESIGN EVALUATION

Mechanical engineering design tasks contain three basic components: (1) determination of a set of needs or conditions to be fulfilled, (2) generation of a set of alternatives, and (3) evaluation of the alternatives. For components manufactured on a large scale, uncertainty as to manufacturing cost can be a significant factor in design evaluation.

Miller and Nevill [1990] use symbolic reasoning for preliminary design. In Dominic, a domain independent iterative redesign system, Dixon and Howe et al. [1986], [Howe, Cohen, et al., 1986] utilize scales to evaluate the current design in the design-evaluate-redesign cycle. Neither of these approaches deal with uncertainty.

### 1.2 HEURISTICS EMBODY ASSUMED USER ATTITUDE TOWARDS UNCERTAINTY

Embedded in the rules of expert systems for mechanical design are not only heuristics for reducing the search space, but also assumptions regarding the user's attitude towards risk and uncertainty. Since these attitudes can vary significantly depending on the manufacturing scenario, the heuristics might not be accurate and/or impose unnecessary constraints on the user. If differences between users are considered at all, the differences are often categorized into stereotypical situations envisioned by the domain expert. Such a system performs well on cases that match the preconceived user profiles, but fail when an atypical user is presented to the system.

### 1.3. RELATED WORK

In dealing with uncertainty, decision makers need to measure and represent uncertainty, combine this information into the decision process, construct a decision model, and draw inferences [Bhatnagar & Kanal, 1986], [Holtzman & Breese, 1986]. While there exist several well known methods for representing uncertainty in rule-base expert systems, no universally accepted generic method for "uncertainty handling" exists [Chandrasekaran & Tanner, 1986]. Wise & Henrion [1986] present a method for comparing several well known methods such as MYCIN's certainty factor, Prospector, Bayes' networks, fuzzy set theory, Dempster-Shafer belief functions, and some non-numerical schemes. Several researchers have noted the advantages of combining formal decision theoretic techniques such as utility analysis [von Neumann & Morgenstern, 1947], [Keeney & Raiffa, 1976] with expert system methods [Keeney, 1986], [Henrion & Cooley, 1987], [Kalagnanam & Henrion, 1990]. Other researchers use utility analysis with expert systems [Sykes & White, 1985], [Spillane & Brown, 1986], [Gabbert & Brown, 1987], but do not include consideration of uncertainty.

## 1.4 COMPARISON BETWEEN UTILITY ANALYSIS AND FUZZY SETS FOR DESIGN EVALUATION

One approach for dealing with this type of uncertainty is fuzzy set analysis [Zadeh, 1975], [Bellman & Zadeh, 1970]. Thurston compares utility and fuzzy set analysis for design evaluation of multiple attributes [Thurston & Carnahan, 1990]. The steps in applying utility analysis and fuzzy set analysis are similar. Each requires the enumeration of relevant attributes, some type of assessment of the relative "value" or "importance" the designer places on each attribute with respect to the other attributes, and the relative attribute performance levels represented by each alternative. Several differences exist:

1. *Quantification of Attributes* - Utility analysis requires that the expected level of performance be quantified, such as dollars, pounds or a numeric scale (such as 1-10). Fuzzy set analysis does not require such quantification, allowing expression in terms of linguistic variables such as "high" and "low."

2. *Monotonicity of Preference* - Fuzzy set analysis permits direct evaluation of attributes whose most desirable level is in the mid-level of the acceptable range, by determining the closeness to a fuzzy goal. Utility analysis requires monotonicity of preference over the attribute range.

3. *Uncertainty* - Both utility and fuzzy set approaches can include consideration of uncertainty as to ultimate performance or attribute levels of a design alternative. Fuzzy sets utilize membership functions, while utility analysis utilizes probability distribution functions to model uncertainty and calculate expected utility.

4. *Relative Importance of Attributes* - Fuzzy set analysis may incorporate "fuzziness" as to the relative "importance" of attributes. Utility analysis does not directly deal with this type of uncertainty, although sensitivity analysis of results on the values of the scaling constants $k_j$ may be performed.

5. *Ordinal Rankings of Alternatives* - Both approaches provide an ordinal ranking of alternatives. However, utility analysis can then be used to quantify beneficial tradeoffs between attributes to guide the iterative design process.

For these reasons, fuzzy sets are more appropriate at the earliest stages of preliminary design configuration when only semantic descriptors of expected design performance are available, and where preliminary design evaluation is being performed by a group, where the use of semantics facilitates reaching consensus. Utility analysis is more appropriate in the later iterative design process that we are concerned with.

## 2. INTEGRATION OF USER-DEFINED EVALUATION FUNCTION INTO EXPERT SYSTEM

A technique to allow designers to incorporate their own unique attitude towards uncertainty as opposed to those assumed by the expert's rules is presented. We describe a method for integrating quantitative procedures for design evaluation which reflect an individual's unique preferences. The general approach is to identify the aspects of heuristic rules which directly or indirectly include assumptions regarding the user's attitude towards risk and replace them with those of the individual end-user. Multiattribute utility analysis is used to methodically extract, interpret, and manage the user's preferences during the construction of the knowledge base. We present the results of the integration of multiattribute utility analysis with a rule-based system for material selection.

### 2.1 ANALYSIS OF HEURISTIC RULE BASE

Figure 1 shows the basic steps. The heuristics used to construct a conventional rule base are analyzed in order to separate subjective from objective rules. *Objective rules* contain the expert's technical judgment regarding feasible alternative design configurations, materials and/or manufacturing processes that would satisfy specified performance requirements. They allow or disallow a design option because of mechanical or structural reasons. They describe or embody universal, constant physical laws. Objective rules originate both from expert design engineers and from texts which describe standard practice. These rules do not typically vary between domain experts as they are composed of factual information.

*Subjective rules* embed assumptions regarding how a particular end-user of the knowledge-based system (KBS) would value a design alternative, including reasonable assumptions as to the user's attitude towards uncertainty. For example, a rule which selects between a traditional versus an innovative new design might reasonably assume that for high volume mass production, long production runs, the financial risk undertaken in committing to a new design whose manufacturing costs are highly uncertain (even if the design offers some advantages) is unacceptable.

These definitions are used as guidelines to separate objective aspects from subjective aspects of rules in the knowledge base. Once the subjective elements of rules are identified and eliminated, expert heuristics are used only to select or eliminate a design configuration based on the technical design requirements.

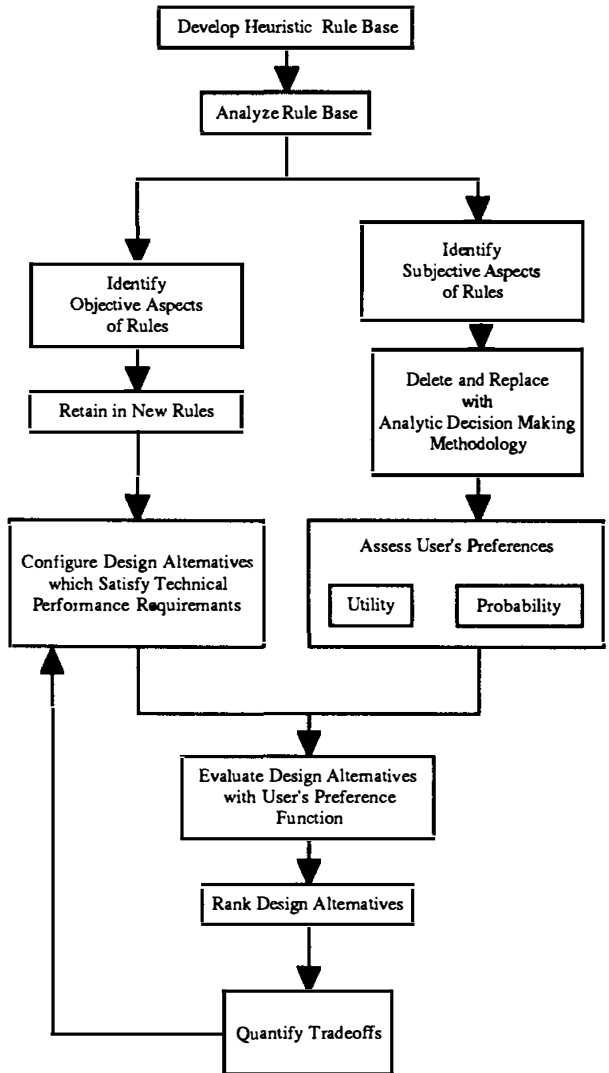

Figure 1. Overview of Steps for Integrating Design Evaluation Procedure into Knowledge Based Systems for Design and Manufacturing

## 2.2 ASSESSING USER'S UTILITY FUNCTION

We have developed a module that assesses the multiattribute utility function of the user through interactive, mouse-driven software. Responses to a sequence of lottery questions (described by Keeney and Raiffa [1976]) determine the single attribute utility functions and scaling constants. The responses reflect the user's attitude towards uncertainty or degree of risk aversion. An outline appears around the selected option, shown in Figure 2.

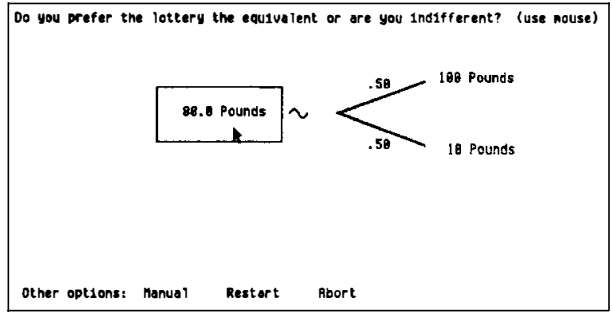

Figure 2. Lottery Question Screen to Determine $u(x_{weight})$

The overall scaling constant K is calculated from equation (2) and the multiplicative multiattribute utility function (after testing for independence conditions) using equation (1).

$$U(X) = \frac{1}{K} [[\prod_{j=1}^{J} (Kk_j U_j(x_j)+1)]-1] \qquad (1)$$

where:

$U(X)$ = overall utility of set of attributes X for each alternative i

$U_j(x_j)$ = single attribute utility function for attribute $x_j$

$x_j$ = performance level for attribute j

$j$ = 1,2...J attributes

$k_j$ = single attribute scaling constant

$K$ = scaling constant, derived from

$$1 + K = \prod_{j=1}^{J} (1 + K k_j) \qquad (2)$$

## 2.3 UNCERTAINTY IN DESIGN PERFORMANCE

The effect of uncertainty as to attribute levels on the desirability of alternatives is reflected by the degree of risk aversion exhibited in the assessed single attribute utility functions. Probabilistic multiattribute utility analysis can be employed to determine the expected value of the overall utility for the $i^{th}$ alternative, $E[U^i(X)]$. It is calculated from expected values of the single attribute utility functions, $E[U_j(x_j)]$; the latter depend on the probability density functions, $f(x_j)$, for the individual attributes. If the attribute levels are independent random variables, it can be shown that

$$E[U^i(X)] = \frac{1}{K}\{[\prod_{j=1}^{J}(Kk_j E[U_j(x_j)]+1)] - 1\} \qquad (3)$$

where

$$E[U_j] = \int_{x_{min}}^{x_{max}} U_j(x_j)f(x_j)dx_j \qquad (4)$$

The expected overall utility, $E[U^i(X)]$, is calculated for each of the I alternatives using equation (3), substituting the expected single attribute utility values in place of their deterministic counterparts.

The uncertainty associated with an attribute is characterized by a probability distribution function. The beta distribution is recommended since it may be readily characterized with input from the design decision maker which is fairly straightforward to assess. The beta distribution is part of the theoretical basis for Project Evaluation and Review Technique (PERT) employed to determine the optimal schedule of inter-dependent tasks with user-estimated uncertainty in completion times [Moder & Phillips, 1970], [Sasieni, 1986]. The required inputs are the minimum, maximum, and most probable values.

A beta random variable distributed on the interval $(x_L, x_U)$ has probability density

$$f(x) = \frac{\Gamma(p+q)}{r\,\Gamma(p)\Gamma(q)}(\frac{x-x_L}{r})^{p-1}(\frac{x_U-x}{r})^{q-1} \quad x_L \leq x \leq x_U$$
$$= 0 \quad \text{otherwise} \qquad (5)$$

where the range is $r = x_L - x_U$. If the shape parameters p and q are chosen to be greater than 1, the distribution is unimodal; if they are equal to one the beta distribution degenerates to a uniform distribution. Of course, the gamma function reduces to the ordinary factorial for an integer argument, i. e., $\Gamma(p) = (p-1)!$

For such a beta variate, the mean $\mu$ and the mode m can be readily calculated by

$$\mu = x_L + r\,\frac{p}{p+q} \qquad (6)$$

and

$$m = x_L + r\,\frac{p-1}{p+q-2} \qquad (7)$$

where the mode, m, is sometimes referred to as the most probable or "most likely" value. Here $x_L$, $x_U$, and m are supplied, so equation 7 defines the relationship between the shape parameters which will produce the requested mode. As p and q vary, however, the probability mass is distributed in a variety of ways about the mode. For instance, when either p or q is large, the density has a pronounced peak near the mode; if both p and q are small, the probability mass is spread over the interval, as shown in Figure 3.

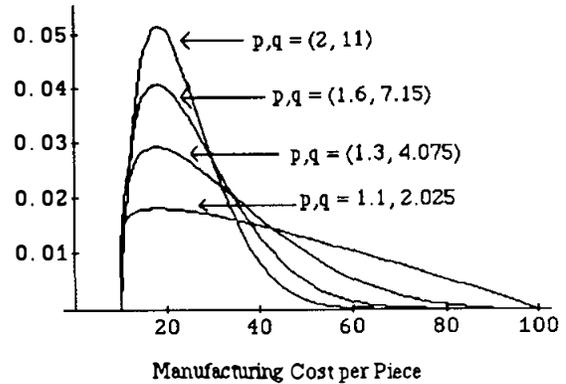

Figure 3. Manufacturing Cost Uncertainty with a Range of Underlying Beta Distribution Parameters p,q

The expected utility for an attribute whose estimated performance level is characterized by a beta probability density function, assuming the interval $(x_L, x_U)$ is contained within $(x_{min}, x_{max})$, and $U_j(x_j) = a-be^{cx}$, from equations 4 and 5 is:

$$E[U_j(x_j)] = \frac{\Gamma(p+q)}{r\,\Gamma(p)\Gamma(q)}\int_{x_L}^{x_U}(a-be^{cx})$$
$$(\frac{x-x_L}{r})^{p-1}(\frac{x_U-x}{r})^{q-1}dx \qquad (8)$$

which, after the change of variables, $y = (x-x_L)/r$,

$$= \frac{\Gamma(p+q)}{\Gamma(p)\Gamma(q)}\int_{0}^{1}[a-be^{c(ry+x_L)}]\,y^{p-1}(1-y)^{q-1}dy \qquad (9)$$

$$= a - \frac{be^{cx_L}\,\Gamma(p+q)}{\Gamma(p)\Gamma(q)}\left\{\sum_{n=1}^{q}(-1)^{n-1}\binom{q-1}{n-1}\left[(-1)^{n+p-1}\frac{(n+p-2)!}{(cr)^{n+p-1}}+\right.\right.$$
$$\left.\left. e^{cr}\sum_{i=0}^{n+-2}(-1)^i\,\frac{i!\binom{n+p-2}{i}}{(cr)^{i+1}}\right]\right\} . \qquad (10)$$

This expression is valid only when the shape parameters p and q are integers greater or equal to 1. Although the expression is somewhat involved, it requires little computational effort when the shape parameters are small, such as less than 10. The expected value can alternatively be obtained by numerical integration of equation 8.

We have written a subroutine which permits the user to provide parameters used to determine a beta distribution which reflects the uncertainty in estimation of performance levels. The input screen is shown in Figure 4.

```
Choose Beta Parameters
Lower bound on variable: ................................. 10
Upper bound on variable: ................................. 100
Explicit value for p : ................................... 1.1
Value for q : ............................................ 2.025
Value of mode : ..........................................
Value of mean : ..........................................

        Abort  [    ]              Do It  [    ]
Find p using mean [    ]   Find p using mode [    ]
```

Figure 4. User Input Screen to Assess Uncertainty via Beta Distribution

## 3. EXAMPLE: AUTOMOTIVE BUMPER MATERIAL SELECTION KBS

Two separate versions of a knowledge-based system (KBS) which performs the task of material selection for an automobile bumper system were developed. The conventional KBS utilizes expert heuristics for all phases of the task. The integrated system substitutes multiattribute utility analysis where rules directly or indirectly deal with uncertainty or evaluate alternative feasible materials.

Bumpers consist of three main components: fascia, bumper beam, and energy absorber (EA), shown in Figure 5. The fascia is an optional outer plastic covering and is made of thermoset or thermoplastic. Vehicles without fascia have either a bright ("chrome") or painted finish.

The energy absorber (EA) allows for energy dissipation without causing permanent damage to the vehicle from low level impacts. The three alternatives are hydraulic strokers, injection molded plastic collapsing column, and expanded bead polypropylene foam. The bumper beam's primary function is to transfer impact energy not dissipated by the energy absorber to the automobile frame. Traditionally beams have been made of steel or aluminum, but reinforced thermosets and unreinforced thermoplastic beams are becoming more common.

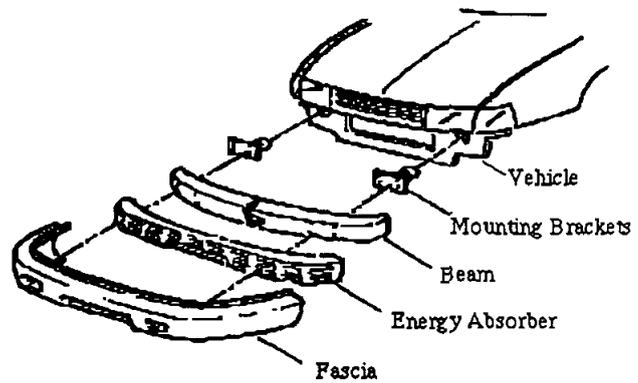

Figure 5. Overview of components of an automobile bumper. [Rusch, 1990]

### 3.1. THE CONVENTIONAL KNOWLEDGE-BASED SYSTEM

Knowledge acquisition came from interviews with bumper design engineers, material manufacturers and from trade and technical publications [Berg, et al., 1989], [Collision Estimating Guide Domestic, 1990], [Delco, 1987], [Delmastro, 1989] and product literature. The knowledge base was constructed in OPS5, running on a Texas Instrument microExplorer$^{TM}$. A mouse-driven menu queries the user for three types of inputs: performance parameters, vehicle characteristics, and manufacturing factors (Figure 6.)

Performance parameters are finish requirements, impact standards, allowable design offset, general bumper shape, and presence or absence of bumper cut-outs. Vehicle characteristics including curb weight, general vehicle model type, and cost range are used to classify the application. Manufacturing data (production volume, length of run, and required lead time) are used both to eliminate materials which could not be manufactured effectively within time requirements, and to estimate manufacturing costs.

```
Choose Variable Values:

Vehicle Type: ................. Sedan Subcompact Sport Pick-up truck
Desired finish: ............... Bright Neutral color Match body color Unknown
Bumper shape: ................. Flat Peaked Curved
Cutouts present? .............. Yes No
Highest allowed offset: ....... Large Medium Small
Cost range: ................... High Medium Low
Impact rating: ................ Over 5mph 5mph 2.5 mph No standard
Curb weight in lbs: ........... 2200
Production volume, x 1000 : ... 193
Length of run in years: ....... 6
Lead time required, in years: . 2
Do It [    ]
```

Figure 6. Conventional KBS Bumper Input Menu

Domain knowledge fell into three categories: configuration rules, design restrictions, and applicability considerations.

*Design configuration rules* restricted certain combinations of materials, since components chosen for one element of the bumper dictate which materials can or cannot be used in another component.

*Design restriction rules* prevented selection of infeasible material alternatives. For example, if a short start-up time was required, materials that could not be manufactured at the indicated volume in the required time would be eliminated.

*Applicability rules* selected a material when more than one material could meet the configuration and design parameter requirements.

## 3.2 THE INTEGRATED SYSTEM

Once the first system was completed and its performance verified, the integrated system was developed. The addition of the probability and utility assessment modules was the only change obvious to the user, while internally the expert system changed significantly. The heuristics used in the original rule base were re-examined to separate subjective from objective rules.

The majority of subjective rules were of the "applicability" type while "design restriction" and "design configuration" rules were primarily objective. The form of the productions in the rule-based system changed from selecting a single material for each component to rejecting materials which did not fall within the stated design parameters and the configuration constraints. Leaving only the objective antecedents in rules resulted in a system that selects a larger number of material alternatives, then ranks these alternatives using the end-user's utility function.

The domain expert's assumptions as to risk aversion characteristics of the user are replaced with the assessed utility function. Domain expert knowledge about possible configurations and minimum performance requirements remains. This allows the revised system to be more responsive to individual user's requirements, especially in instances where the end-user's preference function differs from the stereotypical profile normally assumed by experts.

## 3.3. COMPARISON BETWEEN CONVENTIONAL KBS AND MAUA/INTEGRATED KBS

A truck designer was characterized in the conventional system as a consistently risk-averse decision maker who was willing to spend very little to achieve higher performance or lower weight. Table 1 compares the results of the conventional KBS with that of the integrated KBS. When the typical truck designer utility and scaling parameters were input into the integrated KBS, the output mirrored the convention KBS: a simple, one piece stamped steel bumper with no fascia and no additional energy absorbing unit.

Method Used to Select Bumper Component Materials

| Component | Conventional KBS | Integrated KBS w/typical user | Integrated KBS w/atypical user | System Found on Vehicle |
|---|---|---|---|---|
| Fascia | None | None | Thermoset | None |
| Energy Absorber | None | None | Foam | None |
| Beam | Stamped Steel | Stamped Steel | Stamped Steel | Stamped Steel |
| Utility of System | Atypical: 0.64 | Typical: 0.80 | Atypical: 0.80 | |

Table 1. Comparison of Conventional and Integrated KBS for a Truck Application

Column 3 indicates integrated KBS results using an "atypical" user utility function. "Atypical" users differ from typical users in their toward uncertainty, which is expressed in the degree of risk-aversion exhibited in the utility function. The atypical user in this example showed a much lower degree of risk aversion than the typical user.

When the integrated system used the atypical user profile with the same design parameters as the "typical" designer in the original system, different materials were selected for the fascia and energy absorber. Because design restrictions did not eliminate any of the fascia materials from consideration, the utility module was free to select the material with the highest utility for the user from a list of all possible materials. When the overall utility of the materials was computed, "thermoset," which has the highest appearance ranking, had the highest utility for the "atypical" user. The addition of a fascia allowed foam and collapsing column to be considered as an energy absorbing material, where for the normal user the absence of a fascia eliminated these options from consideration. The energy absorber selected was foam. While no impact standards had to be met, the user still preferred higher impact performance to lower performance. The foam energy absorber provides higher performance than simple fascia support brackets (no EA) at a similar cost.

Both the typical and atypical user had similar utilities for the beam requirements, and the material choice remained the same – stamped steel. The last row in Table 1 indicates that for an atypical user, the overall utility of the system recommended by the integrated KBS, 0.80, is greater than that recommended by the conventional KBS, 0.64, indicating that the integrated system led to a superior alternative.

## 4. CONCLUSIONS

We have shown that heuristics which embody reasonable assumptions regarding the user's attitude towards risk and uncertainty in evaluating alternatives might be inaccurate for some users.

A tool for assessing the user's attitude and a technique for integrating it into the rule base has been presented. The integration is performed at the individual rule level, and not simply tacked onto the end of the expert system. The heuristic rule base is analyzed, making a clear distinction between aspects of rules which reflect objective technical expertise and aspects which include assumptions as to the user's attitude towards uncertainty.

These attitudes can be successfully dealt with in a more direct manner by eliminating the domain expert's assumptions regarding the user's preferences and replacing them with multiattribute utility analysis. Expert heuristics still play a major role in generating feasible design alternatives from a purely technical viewpoint.

The example showed that this integrated approach can lead to improved selections for the atypical user, without compromising system performance for the more stereotypical designer. Sensitivity to users' preferences is provided without disturbing the original expert's objective knowledge.

An additional benefit is that incorporating new knowledge regarding technological advances is simplified. The only productions that need to be added are configuration restrictions and minimum performance levels, while uncertainty considerations are reflected in the utility and probability assessment modules.

By assessing the utility and probability functions of design engineers and directly incorporating them into the rule base, a computer aid to design has been developed which permits engineers to develop designs which are optimal for their own decision making environment.

### Acknowledgment

The author gratefully acknowledges the support of the National Science Foundation under grant DMC-8809829 and PYI award DDM-8957420.

### References

Bellman R. E. and Zadeh, L. A., "Decision-Making in a Fuzzy Environment," *Management Science*, Vol. 17B, 1970, pp. 141 - 164.

Berg, J. W., R. E. Morgan, and G. A. Klumb "High Performance RIM Fasica," *Plastics in Automobiles, International Congress and Exposition, Detroit, Michigan, 1989.* p. 33.

Bhatnagar, R. K., Laveen N. Kanal, Handing Uncertain Information: A Review of Numeric and Non-numeric Methods, *Uncertainty in Artificial Intelligence,* Kanal, L.N. et al. (Eds.), Elsevier Science Publishers B. V., 1986, p. 3-26.

Chandrasekaran, B., Michael C. Tanner, Uncertainty Handling in Expert Systems: Uniform vs. Task-Specific formalisms, *Uncertainty in Artificial Intelligence,* Kanal, L.N. et al. (Eds.), Elsevier Science Publishers B. V., 1986, p. 35-46.

Collision Estimating Guide Domestic. Mitchell International, April, 1990.

Delco Products Division of General Motors Corp. *Hard Bar Bumper System Design Manual,* Dec. 14, 1987.

Delmastro, J. "Overview of the North American Car Market for Bumper Applications," *Plastics in Automobiles, International Congress and Exposition, Detroit, Michigan, 1989.* pp. 227-234.

Dixon, J.R., A. Howe, P.R. Cohen, and M.K. Simmons, "Dominic I: Progress Towards Domain Independence in Design by Iterative Redesign," *Proceedings of the ASME 1986 Computers in Engineering Conference, Chicago, Illinois, July, 1986,* Vol I, pp. 199-212.

Gabbert, P., D.E. Brown, A Knowledge-Based Approach to Materials Handling System, *World Probuctivity Forum & 1987 International Industrial Engineering Conference,* May 17-20, 1987, p.445-451.

Henrion, M., D. R. Cooley, "An Experimental Comparison of Knowledge Engineering for Expert Systems and for Decision Analysis," *Proceedings of AAAI-87,* Seattle, WA. 1987, p. 471-476.

Holtzman, S., J. Breese, "Exact Reasoning about Uncertainty: On the Design of Expert Systems for Decision Support," *Uncertainty in Artificial Intelligence,* Kanal, L.N. et al. (Eds.), Elsevier Science Publishers B. V., 1986, p. 339-345.

Howe, A., P. Cohen, J. Dixon, and M. Simmons, "Dominic: A Domain-Independent Program for Mechanical Engineering Design," *Applications of Artificial Intelligence in Engineering Problems Conference, Southampton University, U.K., April, 1986. Vol. 1.* pp. 290-300.

Kalagnanam, Jayant, Max Henrion, A Comparison of Decision Analysis and Expert Rules for Sequential

Diagnosis, *Uncertainty in Artificial Intelligence 4*, R. D. Shachter et al. (Eds.), Elsevier Science Publishers B. V., 1990, p.271-281.

Keeney, R.L. and H. Raiffa, *Decisions with Multiple Objectives: Preferences and Value Tradeoffs*. John Wiley & Sons, Inc., 1976.

Keeney, R. L., Value-Driven Expert Systems for Decision Support, Expert Judgment and Expert Systems, Mnmpower, J.L. et al. (Eds.), 1986, p. 155-171.

Moder, J., and Phillips, C., Project Management with CPM and PERT, Van Nostran Reinhold Co., New York, 1970.

Miller, V.T., Nevill, G.E., "Knowledge Sifting" for Preliminary Design," *Proceedings of the Design Theory and Methodology Conference*, 1990, ASME.

Rusch, K. C. "Overview of Automotive Plastic Bumpers," *Automobile Bumper Systems and Exterior Panels*, 1990, SAE paper 900420.

Sasieni, M., "A Note on PERT Times," *Management Science*, 32, 12, December, 1986.

Spillane, A. R., D. E. Brown, An Intelligent Design Aid for Large Scale Systems with Quantity Discount Pricing, *Proceedings of the 1986 IEEE International Conference in Systems, Man, and Cybernetics*, Oct. 14-17, 1986, p. 1331-1336.

Sykes, E. A., C. C. White, III, Specifications of a Knowledge System for Packet-Switched Data Network Topological Design, *IEEE Expert Systems in Government Symposium*, 1985, p. 102-110.

Thurston, D. L. and Carnahan, J. V., "Fuzzy Ratings and Utility Analysis in Preliminary Design Evaluation of Multiple Attributes," University of Illinois Technical Report, Department of General Engineering, 8/90.

Thurston, D.L., "Multiattribute Utility Analysis in Design Management," *IEEE Transaction on Engineering Management*, Vol. 37, No. 4, November, 1990.

von Neumann, J. and O. Morgenstern, Theory of Games and Economic Behavior, 2nd ed. Princeton University Press, Princeton, N.J., 1947.

Wise, B. P., Max Henrion, A Framework for Comparing Uncertain Inference Systems to Probability, *Uncertainty in Artificial Intelligence*, Kanal, L.N. et al. (Eds.), Elsevier Science Publishers B. V., 1986, p. 69-83.

Zadeh, L. A., "The Concept of a Linguistic Variable and its Application to Approximate Reasoning," *Information Sciences*, Vol. 8, 1975, pp. 199-249.